# ACQ: Improving Generative Data-free Quantization Via Attention Correction

**Jixing Li [a,b], Xiaozhou Guo [a,b], Benzhe Dai [a,b], Guoliang Gong [a,b,c], Min Jin [a,b,c], Gang Chen [a,b,c*], Wenyu Mao [a,b,c], Huaxiang Lu [a,b,c,d,e]**

[a]*Institute of Semiconductors, Chinese Academy of Sciences, Beijing 100083, China*

[b]*University of Chinese Academy of Sciences, Beijing 100049, China*

[c]*Semiconductor Neural Network Intelligent Perception and Computing Technology Beijing Key Laboratory, Beijing 100083, China*

[d]*Materials and Optoelectronics Research Center, University of Chinese Academy of Sciences, Beijing 100049, China*

[e]*College of Microelectronics, University of Chinese Academy of Sciences, Beijing 100049, China*

∗ Corresponding author at: Institute of Semiconductors, Chinese Academy of Sciences, Beijing 100083, China

E-mail: chengang08@semi.ac.cn (Guoliang Gong)

## Abstract

Data-free quantization aims to achieve model quantization without accessing any authentic sample. It is significant in an application-oriented context involving data privacy. Converting noise vectors into synthetic samples through a generator is a popular data-free quantization method, which is called generative data-free quantization. However, there is a difference in attention between synthetic samples and authentic samples. This is always ignored and restricts the quantization performance. First, since synthetic samples of the same class are prone to have homogenous attention, the quantized network can only learn limited modes of attention. Second, synthetic samples in eval mode and training mode exhibit different attention. Hence, the batch-normalization statistics matching tends to be inaccurate. ACQ is proposed in this paper to fix the attention of synthetic samples. An attention center position-condition generator is established regarding the homogenization of intra-class attention. Restricted by the attention center matching loss, the attention center position is treated as the generator's condition input to guide synthetic samples in obtaining diverse attention. Moreover, we design adversarial loss of paired synthetic samples under the same condition to prevent the generator from paying overmuch attention to the condition, which may result in mode collapse. To improve the attention similarity of synthetic samples in different network modes, we introduce a consistency penalty to guarantee accurate BN statistics matching. The experimental results demonstrate that ACQ effectively improves the attention problems of synthetic samples. Under various training settings, ACQ achieves the best quantization performance. For the 4-bit quantization of Resnet18 and Resnet50, ACQ reaches 67.55% and 72.23% accuracy, respectively.

*Keywords:* generative data-free quantization; homogenous attention; attention center matching; adversarial loss; consistency penalty

## 1. Introduction

In recent years, neural networks have achieved great success in many applications, such as image recognition [1], object detection [2], and semantic segmentation [3]. Nevertheless, it is challenging to deploy a neural network in a resource-constrained device due to the enormous memory usage and complicated computation of the neural network. This issue promotes the development of model compression and acceleration techniques, including knowledge distillation [4], lightweight network design [5], as well as pruning [6]. Network quantization represents the floating-point parameters and activations within the networks by low-bit integers. Since more and more hardware (such as GPU, FPGA, and AI chips) support low-bit calculation, neural network quantization has emerged as a popular model compression and acceleration technique.

The rounding and truncation operations can bring quantization noise to the original network in quantization calculation. This may weaken the performance of the quantized network to a certain extent. Under ultralow bit width, some quantized networks may even be in complete failure. When the training data is accessible, quantization-aware training (QAT) [7] is proven to be a technique that can offer desirable accuracy. This method can reduce the quantization noise by adjusting the weight through backpropagation. Nevertheless, the dependence on training data also forms a defect of this method since the original training data (such as medical records, confidential commercial pictures, or military pictures) are inaccessible in many scenes for privacy and security reasons.

Given the above considerations, data-free quantization has attracted much attention from researchers [8-10]. This technique can quantize a network precisely without accessing any authentic data. It is often assumed in the early DFQ [10] and ACIQ [11] that the neural network activation satisfies normal or Laplace distribution. They focused on solving quantization coefficients and realized comparable 8-bit quantization precision. However, these methods cannot be adopted for ultralow-bit quantization because of their high precision loss. Some researchers obtained synthetic samples that resemble the distribution of the authentic sample by using information from the pre-trained full-precision network, such as batch-normalization (BN) statistics. These approaches can be categorized into noise-optimized data-free quantization [8, 12-14] and generative data-free quantization [9, 15-17]. The former initializes a sample that satisfies the gaussian distribution, and the dimension of the sample is consistent with the size of a real sample. It is then optimized by gradient descent. In the latter category, a generative network [18] is employed to convert noise vectors into synthetic samples. Remarkably, each sample in the noise-optimized data-free quantization is created over a thousand times gradient iteration. There is a considerable time cost to generate a large number of samples. This method is mainly used in post-training quantization (PTQ), which requires a few samples [19]. Under ultralow bit width, a clear difference between the performances of PTQ and QAT exists. By contrast, synthetic samples in generative data-free quantization can be generated fast through forward inference. This makes generative data-free quantization the most popular method for ultralow-bit data-free quantization.

However, a significant gap still exists between generative data-free quantized models and quantized models trained with authentic data. This motivated us to explore the difference between synthetic samples and authentic samples. We try to observe them from the pre-trained model's perspective and find their differences in attention [20-22]. We summarized this difference as the homogenization of intra-class attention and attention differentiation in different network modes.

Regarding the homogenization of intra-class attention, the synthetic samples in the same class are shown to have highly similar attention maps. In Fig. 1, we display the same class of synthetic samples of three generative data-free quantization methods (including GDFQ [15], ARC [16], and Qimera [23]) and their attention maps. As you can see, the attention is almost consistent across synthetic samples of the same class. This indicates that the characteristics of synthetic samples were largely in the same positions. For authentic samples, the attention changes as the target objects appear in different positions. The quantized network can only learn limited attention modes due to the intra-class attention homogenization of synthetic samples. This led to the performance restriction.

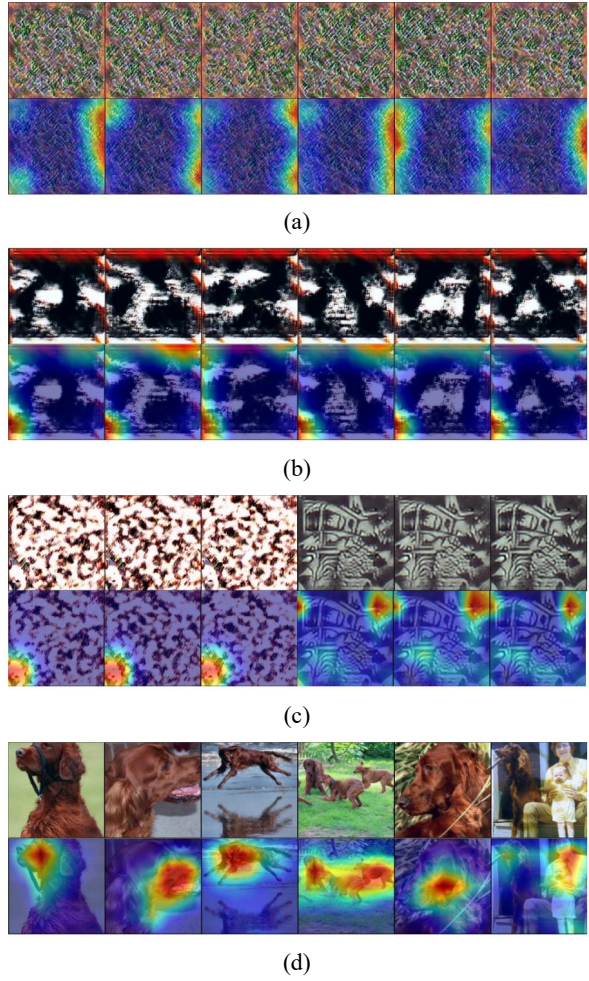

(a)

(b)

(c)

(d)

**Fig. 1.** The same class of authentic samples from Imagenet2012 dataset and GDFQ, ARC, and Qimera synthetic samples generated with Resnet18 as well as their attention maps. (a) GDFQ synthetic samples and their attention maps; (b) ARC synthetic samples and their attention maps; (c) Qimera synthetic samples and their attention maps (the three pictures on the right-hand side are the boundary support samples); (d) authentic samples and their attention maps.

Regarding the attention differentiation in different network modes, it is revealed that the same batch (batch size>= 16) of synthetic samples have highly different attention in training and eval modes. Here, the training mode and eval mode are in terms of forward inference of the pre-trained model. They are mainly distinguished by the calculation mechanism in the BN layer [24]. We show the attention maps of GDFQ synthetic samples in the two modes in Fig. 2. The attention of synthetic samples varies enormously when the network mode changes. However, for authentic samples, the attention just changes slightly. To explore the reason, we select 960 authentic samples and synthetic samples, respectively, that could be identified by a floating point (FP) network in eval mode accurately. These samples are input to the FP network in batches with batch size 16 to obtain the results in Table 1. BNS error indicates the average error between the BN statistics of the input samples and the corresponding stored BN statistics. We observe that the recognition precision of the network in training mode was reduced by 63.5% for synthetic samples, whereas the precision was reduced by 3.8% merely for authentic samples. This implies that along with the change in attention, there are also significant differences in category judgments. More importantly, in training mode, the BNS error of synthetic samples increased significantly, whereas the BNS error of authentic samples decreased. Hence, correct BN statistics matching cannot be achieved for synthetic samples. There exists a clear deviation in BN statistical distribution between synthetic and authentic samples.

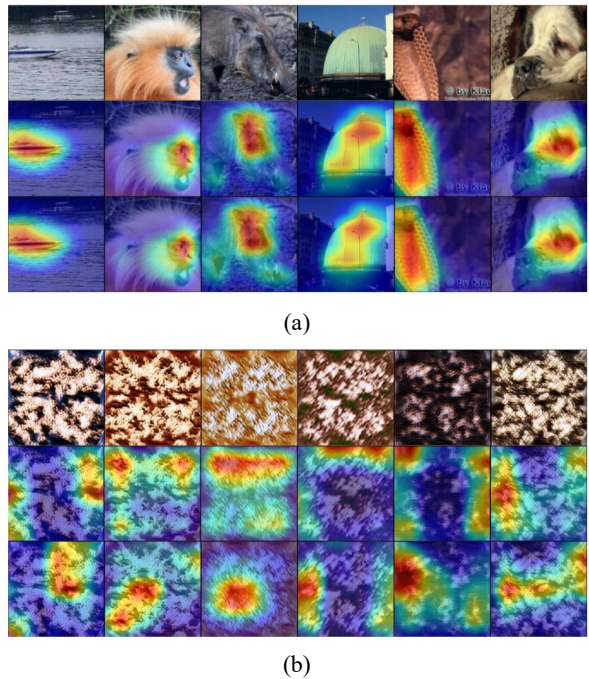

**Fig. 2.** Attention maps of authentic samples from Imagenet2012 dataset and GDFQ synthetic samples generated with Resnet18 in eval and training modes, respectively. The second row of pictures shows the attention maps in eval mode, whereas the third row of pictures displays the attention maps in training mode. (a) Authentic samples and their attention maps; (b) GDFQ synthetic samples and their attention maps.

**Table 1**

The recognition accuracy and BNS error of 960 authentic samples and synthetic samples on Resnet18 in training mode and eval mode.

|  | Authentic sample | Synthetic sample |
| --- | --- | --- |
| Acc (eval mode) | 100% | 100% |
| Acc (training mode) | 96.2% | 36.5% |
| BNS error (eval mode) | 0.61 | 0.92 |

| BNS error (training mode) | 0.47 | 2.76 |
| --- | --- | --- |

In this paper, we come up with an ACQ method, where the homogenization of intra-class attention and the attention differentiation in different network modes of synthetic samples are corrected to improve the performance of the generative data-free quantization. For the homogenization of intra-class attention, we treat the attention center position as the generator's condition input to guide the attention of synthetic samples in coarse grains to cope with the target position to achieve diversity of attention. We design an adversarial loss of paired synthetic samples under the same condition to prevent the generator from paying too much attention to the condition, which could yield mode collapse. Due to the conditional consistency, the generator is induced to pay more attention to the noise vectors. The adversarial loss also encourages the generator to create samples with a great difference for similar noise vectors to boost the diversity of synthetic samples. Regarding attention differentiation in different network modes, a consistency penalty is designed to constrain the BN statistic matching, class judgment, and attention of synthetic samples in different network modes. Therefore, the BN statistical distribution of synthetic samples is closer to that of authentic samples. Our experimental results verify that ACQ has better performance under a variety of quantization training strategies compared with other data-free quantization methods. The contribution of this paper is summarized in the following:

• Employing the attention of authentic samples, we revealed two attention abnormalities from the existing synthetic samples, including the homogenization of intra-class attention and attention differentiation in different network modes.

• We proposed a coarse-grained attention center position-condition generator to resist the homogenization of intra-class attention. Moreover, we designed adversarial loss of paired synthetic samples under the same condition to weaken the generator's attention to the condition.

• We put forward a consistency penalty for accurate BN statistics matching, which improved the attention and class consistency of synthetic samples in different network modes.

• We verify the outstanding performance of ACQ through extensive experiments using the datasets of ImageNet2012, CIFAR10, and CIFAR100.

## 2. Related work

We provide a brief review of data-driven quantization and data-free quantization below. More details of model quantization can be found in the paper [25].

Data-driven quantization: Both QAT and PTQ require real data to complete quantization. By adapting to quantization operation through relearning the weights, the QAT leads to a model quantized to an ultralow bit width maintaining its performance, although it relies on complete training datasets. The main research contents of QAT contain the quantizer design [26], training strategy [27], approximate gradient [28], and binary network [29]. In contrast, PTQ realizes the model quantization employing quite limited data. The 8-bit quantization precision is guaranteed through techniques like scale optimization [30], correcting biases [31], and weight adjustment [32]. In particular, AdaRound [19] provided a self-adaption rounding technique, enhancing the performance of ultralow-bit PTQ significantly. BrecQ [33] reported a network block output-based reconstruction technique. This was based on the theoretical study and the empirical assumption of second-order loss to boost the precision of trained quantization.

Data-free quantization: This method quantizes a network without accessing any authentic data. An assumption the early data-free quantization relied on is that the neural network activation satisfied the normal or Laplace distribution and that it specialized in calibrating the network parameters. Recent studies focused on synthetic samples to enhance the quantization precision. ZeroQ [8] was viewed as the first noise-optimized data-free quantization method. In this method, synthetic samples are created based on the BN statistics matching, which obtains a satisfying result under 8-bit width. DSG [12] further suggested relaxing the batch-normalization statistics alignment to generate more diverse samples. In the IntraQ [14], it was shown that the intra-class diversity of synthetic samples could be strengthened by random cutting and spaced intra-class penalty. Different from the noise-optimized method, a basic framework was constructed in the GDFQ for generative data-free quantization. The data-free quantization performance under ultralow-bit width was improved massively by synthesizing samples with out-of-class intervals and QAT. Based on this, ARC suggested a better generator found by neural architecture search. This improved the quality of synthetic samples with high spatial dimensions. Inspired by adversarial knowledge distillation, ZAQ [17] adopted adversarial training of the generator. This drove the generator to generate difficult samples against the quantized model, which enhanced the capacity of the quantized model. Enlightened by BSS [34], Qimera enabled the generator to synthesize boundary support samples through superposed latent embeddings, which improved the ability of the quantized model to identify the class boundary. In particular, AIT [9] paid no attention to the quality of synthetic samples, but it improved the training strategy of the quantized model.

The previous studies have improved the precision of generative data-free quantization in many aspects. However, they overlook the attention difference between synthetic and authentic samples. We believe that this difference hinders the performance of generative data-free quantization. This paper aims to reduce this gap to achieve advanced performance in generative data-free quantization.

## 3. Method

In this section, we first introduce the ACQ-related knowledge, including quantization, attention map, and the train and eval modes of a neural network. We also briefly review the GDFQ as the general framework of generative data-free quantization. The way the attention of synthetic samples is corrected by ACQ is then described in detail. Finally, we present the training flow of the ACQ.

### 3.1. Preliminaries

#### 3.1.1. Quantizer

In line with the previous works [15, 16, 23], we use the asymmetric uniform quantizer to implement network quantization. Given the weight or activation $x$, the quantization bit width $n$, the lower bound $l$ and upper bound $u$ of the floating-point number, we defined the quantizer as below:

$$\boldsymbol{q} = clip\left(round\left(\frac{\boldsymbol{x}}{s} - b\right), \, -2^{n-1}, 2^{n-1} - 1\right) \#(1)$$

where $clip(\boldsymbol{x}, -2^{n-1}, 2^{n-1} - 1) = \min\left(\max\left(\boldsymbol{x}, -2^{n-1}\right), 2^{n-1} - 1\right)$ and $round(\cdot)$ represents the rounding to the nearest neighbor. $s$ and b can be computed by $s = \frac{u-l}{2^n - 1}$ and $b = \frac{l}{s} + 2^{n-1}$. We define the corresponding fake quantized data $\boldsymbol{x_{fake}}$ as follows:

$$\boldsymbol{x_{fake}} = (\boldsymbol{q} + b) \cdot s \#(2)$$

For activations and weights, we use a layer-wise quantizer and a channel-wise quantizer, respectively. The lower bound $l$ and upper bound $u$ are set to the minimum and maximum of per-layer activations (per-channel weights). The quantizer is embedded in the floating-point network. Network weights are adjusted by using backpropagation in the QAT to reduce the quantization error.

#### 3.1.2. Attention map

The attention map can be employed to represent the attention of a neural network to varied areas of the input picture. The target object is often located in the high-response area of the attention map. Many attention map calculation methods have been reported, which include GradCAM [20] and ScoreCAM [21]. In this paper, we adopt the attention map calculation method developed by Sergey [22]. Given the backbone output $A \in \mathbb{R}^{c \times h \times w}$, the attention matrix $M \in \mathbb{R}^{h \times w}$ is defined as:

$$M = norm\left(\sum_{i=1}^{C} |A_i|^2\right) \#(3)$$

where $c$, $h$, and $w$ represent the number of channels, height, and width of the backbone output feature, respectively; $A_i = A(i, :, :)$ (using Matlab notion); $norm(\cdot)$ represents a linear normalization function. We obtained the attention map when the matrix $M$ was zoomed to the same dimensions as the input picture.

### 3.1.3. Training mode and eval mode

Regarding the forward inference of a pre-trained model, the main difference between the train and eval modes is the calculation mechanism within the BN layer:

$BNLayer(x) = \gamma \frac{x-\mu}{\sigma} + \beta \#(4)$ In the training mode, the calculation in the BN layer was performed employing the average value and variance of the input feature $x$, namely $\mu = mean(x), \sigma = std(x)$. In the eval mode, $\mu$ and $\sigma$ adopt the pre-stored BN statistics parameters obtained from the average value and variance of the input feature, which was renewed by sliding in training. To some extent, the two parameters could be viewed as the statistical parameters of an authentic sample distributed in this network layer. Hence, the network's attention map and class judgment on the samples are almost the same in train and eval modes when a batch of authentic samples was input into the pre-trained model.

The attention differentiation of synthetic samples in different network modes reveals an obvious error between the BN statistics of synthetic samples and authentic samples. This indicates that there is some difference between the statistical distribution of synthetic and authentic samples. Typically, in the training mode, the forward inference of the neural network will update the BN statistics parameters. This, nevertheless, is not allowed in our work as the FP network is pre-trained.

### 3.1.4. GDFQ

We use the most universal GDFQ as the basic framework. In Fig. 3, we show that for the FP model $F^{eval}$ and its quantized model $Q$, GDFQ aims to construct the synthetic data $(\hat{x}, y)$ and recover the precision of the quantized model by using QAT. $F^{eval}$ means that the FP model is in eval mode and $F^{train}$ means that the FP model is in training mode.

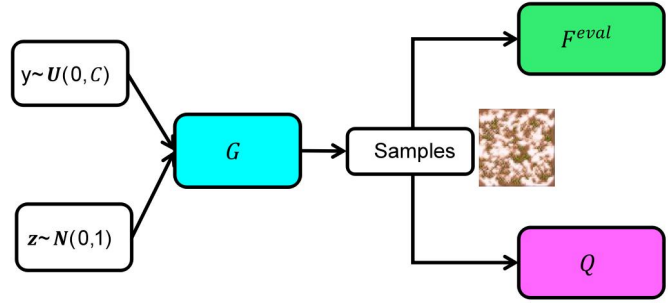

**Fig. 3.** Framework of the GDFQ.

Specifically, the condition generator $\boldsymbol{G}$ converted the class label $y \sim \boldsymbol{U}(0, C)$ and noise vector $\boldsymbol{z} \sim \boldsymbol{N}(0,1)$ into a synthetic sample $\hat{\boldsymbol{x}}$. Here, $C$ represents the total number of classes, $\boldsymbol{U}$ represents the uniform distribution, and $\boldsymbol{N}$ represents the Gaussian distribution:

$$\hat{\boldsymbol{x}} = \boldsymbol{G}(\boldsymbol{z}|\boldsymbol{y}) \#(5)$$

The loss of the generator consisted of the BN statistic matching loss function $L_{BNS}^{GDFQ}(\boldsymbol{G})$ and the classification loss function $L_{CE}^{GDFQ}(\boldsymbol{G})$. $L_{BNS}^{GDFQ}(\boldsymbol{G})$ aligns the BN statistic of synthetic samples with the BN statistical stored in the BN layer of the full-precision model. This helps the distribution of synthetic samples approximate the authentic distribution. $L_{CE}^{GDFQ}(\boldsymbol{G})$ means the cross-entropy loss, which guides the generator in synthesizing samples with heterogeneous spacing.

$$L_{BNS}^{GDFQ}(\boldsymbol{G}) = \sum_{l=1}^{L} \left\| \boldsymbol{\mu}_l^{s,eval} - \boldsymbol{\mu}_l \right\|_2^2 + \left\| \boldsymbol{\sigma}_l^{s,eval} - \boldsymbol{\sigma}_l \right\|_2^2 \#$$
$$\#(6)$$

$$L_{CE}^{GDFQ}(\boldsymbol{G}) = E_{\boldsymbol{z},y}\left( CE\big(\boldsymbol{F}^{eval}\big(\boldsymbol{G}(\boldsymbol{z}|y)\big), y\big) \right) \#(7)$$

where $\boldsymbol{\mu}_l^{s,eval}$ and $\boldsymbol{\sigma}_l^{s,eval}$ are the average value and variance of synthetic samples in the $l$-th BN layer of the FP network in eval mode, respectively; $\boldsymbol{\mu}_l$ and $\boldsymbol{\sigma}_l$ represent the average value and variance of the $\boldsymbol{F}^{eval}$ stored in the $l$-th BN layer, respectively.

The loss functions of the quantized model include the classification loss function $L_{CE}^{GDFQ}(\boldsymbol{Q})$ involving the class label $y$ and the KL distance loss function $L_{KD}^{GDFQ}(\boldsymbol{Q})$ involving class probability distribution. The former function expected that the quantized model could classify synthetic samples as accurately as the FP model. The latter function stemmed from knowledge distillation. In other words, the output of the quantized model should be close to that of the FP model enough to guarantee that the performance of the quantized model was comparable to that of the FP model considering the same synthetic sample.

$$L_{CE}^{GDFQ}(\boldsymbol{Q}) = E_{\widehat{x},y}\big(CE(\boldsymbol{Q}(\widehat{x}),y)\big) \#(8)$$

$$L_{KD}^{GDFQ}(\boldsymbol{Q}) = E_{\widehat{x}}\left(KL\big(\boldsymbol{Q}(\widehat{x}),F^{eval}(\widehat{x})\big)\right) \#(9)$$

## 3.2. ACQ

We proposed ACQ to enable the attention of synthetic samples approaching authentic samples. The homogenization of intra-class attention and the attention differentiation in different network modes of synthetic samples will be improved. In Fig. 4, we show the generator training framework of the ACQ. The framework consists of a generator $\boldsymbol{G}$, an FP network $\boldsymbol{F^{eval}}$, and an FP network $\boldsymbol{F^{train}}$. We regard the classification loss and BN statistics matching loss of the GDFQ as the loss functions of the generator to guide the generator in learning the class boundary and BN statistical distribution. Moreover, we treat the attention center position label as the condition input of the generator to boost the diversity of intra-class attention. The attention of synthetic samples is led by the position label to focus on the corresponding target position under the guidance of coarse-grained attention center matching loss. In particular, the paired samples are generated with identical input conditions by the generator at the same time. The adversarial loss is designed to maximize the difference between paired samples, preventing the generator from paying too much attention to the condition to yield mode collapse. $\boldsymbol{F^{train}}$ poses a consistency penalty to the loss functions, including the classification loss, BN statistics matching loss, and coarse-grained attention center matching loss. The purpose is to make the BN distribution of synthetic samples tend to the BN distribution of authentic samples. This improved the consistency in attention of synthetic samples in different network modes. We show the relevant details below.

### 3.2.1. Coarse-grained attention center match

As mentioned above, the attention map of a neural network is determined by an attention matrix $\boldsymbol{M}$. The extent of the network's attention to the receptive area of the picture is reflected by the matrix element ($\boldsymbol{M}(i,j) \in [0,1]$). To enhance the diversity of intra-class attention, we treated the attention matrix as the condition input of the generator to control the attention of synthetic samples. Nevertheless, it is challenging to artificially design an attention matrix a priori. There is no definite mathematic model that could model an attention matrix.

Fortunately, the maximum value of the attention matrix is unique, and its position $p$ represents the attention center, which is the most significant feature area. Thus, this paper did not seek fine-grained control of the attention matrix but aimed to obtain diverse attention centers, which would realize coarse-grained attention center matching. This practice enjoys two advantages. First, there is no need to design an attention matrix a priori but input the attention center position $p \sim \boldsymbol{U}(0, hw)$ as the condition. The condition can be simplified, which makes the generator easy to converge. Second, the attention to other positions is overlooked. This provides a vast exploration space for the generator and further intensifies the attention to diversity.

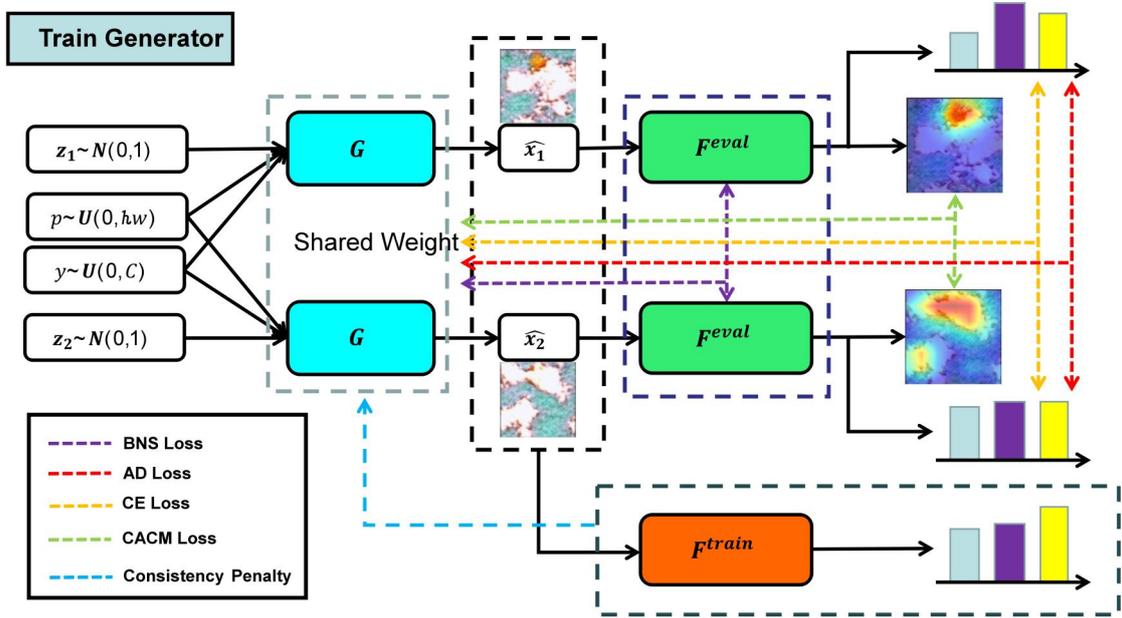

**Fig. 4.** Generator training framework of the ACQ.

We regard the attention matrix as a probability matrix for determining whether a position is the attention center . We simplify the attention center matching as a binary classification. Given the position label of an attention center, it is expected that the attention matrix of synthetic samples at this position was as large as possible. We define the coarse-grained attention center matching loss function as below:

$$L_{CACM}^{ACQ}(\boldsymbol{G}) = BCE\left(\max\left(\boldsymbol{M}\left(\frac{p}{h}, p\%w\right) + \varepsilon_1, 1\right), 1\right) \#$$
$$\#(10)$$

where $\epsilon_1$ represents the relaxing factor for preventing the $\mathrm{M}\left(\frac{p}{\mathrm{H}}, p\%\mathrm{W}\right)$ from being over small in the training and causing gradient explosion.

ACQ adopts a generator structure in line with the GDFQ. A slight adjustment is made to blend the position condition label $p$. Specifically, the position condition label $p$ and class condition label $y$ can be blended with the noise vector $\boldsymbol{z}$ directly for low spatial dimensions of datasets such as CIFAR-10 and CIFAR100. This would form the input vector $\boldsymbol{i}$:

$$\boldsymbol{i} = (Embedding_{class}(y) + \boldsymbol{z}) \cdot Embedding_{position}(p) \#(11)$$

where $Embedding_{class}(\cdot)$ and $Embedding_{position}(\cdot)$ represent learnable encoding layers encoding the class label and position label into class and position vectors, respectively. It would be useful to blend the position condition label with the medium features of the generator for datasets with high spatial dimensions, such as ImageNet, to facilitate the convergence of the generator. As shown in Fig. 5, we convert the position label $p$ into a position vector by one-hot encoding and label smoothing first and then adjust it into a position matrix $\boldsymbol{P} \in \mathbb{R}^{h \times w}$ in the same dimensions as the attention matrix. The input noise vector is converted into a medium feature $\boldsymbol{f}$ through a fully-connected network. Following that, the $\boldsymbol{f}$ is downsampled by adaptive maximum pooling and adaptive average pooling, obtaining the features $\boldsymbol{f_1} \in \mathbb{R}^{c_1 \times h \times w}$ and $\boldsymbol{f_2} \in \mathbb{R}^{c_2 \times h \times w}$. Employing a convolutional network, we blend the position matrix P with the features $\boldsymbol{f_1}$ and $\boldsymbol{f_2}$, forming the feature $\boldsymbol{f}'$. Finally, the feature $\boldsymbol{f}'$ is upsampled and added with $\boldsymbol{f}$, which yields the feature $\boldsymbol{\mathcal{F}}$ blended with the position label. We describe this process below:

$$\boldsymbol{P} = label\_smooth(one\_hot(p))$$

$$\boldsymbol{f} = Linear(\boldsymbol{z})$$

$$\boldsymbol{f_1} = AdaMaxpool(\boldsymbol{f})$$

$$\boldsymbol{f_2} = AdaAvgpool(\boldsymbol{f})$$

$$\boldsymbol{f}' = conv([\boldsymbol{f_1}, \boldsymbol{f_2}, \boldsymbol{P}])$$

$$\boldsymbol{\mathcal{F}} = Up(\boldsymbol{f}') + \boldsymbol{f} \#(12)$$

Employing the proposed generator structure and coarse-grained attention center matching loss, the generator could generate corresponding synthetic samples as per the given attention center position labels. Moreover, since the constraint was coarse-grained, some attention differences between synthetic samples existed even with identical attention center position labels. Accordingly, we improved the homogenization of intra-class attention of synthetic samples.

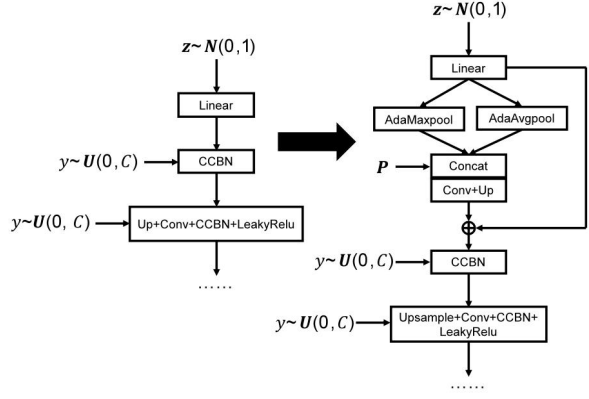

**Fig. 5.** The generator for synthesizing samples with high spatial dimensions. CCBN represents the class condition BN layer [35]. The left part shows the generator structure of the GDFQ and the right part shows the adjustment for blending the position labels to the ACQ.

### 3.2.2. Adversarial loss

Since a condition is often structured, the generator is prone to pay attention to the condition and vector. Therefore, a conditional generative network is convenient for triggering a mode collapse.

In ACQ, the attention center position condition further boosts the probability of mode collapse. Hence, we design an adversarial loss of paired samples under identical conditions. We show in Fig. 4 that noise vectors $z_1$ and $z_2$ are sampled from the Gaussian distribution. Using the identical class label $y$ and attention center position label $p$, we derive two synthetic samples $\hat{x}_1 = G(z_1|y,p)$ and $\hat{x}_2 = G(z_2|y,p)$ as well as their classes $o_1$ and $o_2$ judged by $F_{eval}$. We defined the adversarial loss as follows:

$$L_{AD}^{ACQ}(G) = \frac{MAE(z_1, z_2)}{JS(y_1, y_2)}$$

$$y_1 = softmax(F_{eval}(\hat{x}_1)^{\neg\{o_1, o_2\}})$$

$$y_2 = softmax(F_{eval}(\hat{x}_2)^{\neg\{o_1, o_2\}})$$

$$JS(y_1, y_2) = \frac{1}{2}(KL\left(y_1, \frac{y_1 + y_2}{2}\right) + KL(y_2, \frac{y_1 + y_2}{2}))$$

$$MAE(z_1, z_2) = mean(|z_1 - z_2|)$$
$$\#(13)$$

The adversarial loss can explicitly minimize the ratio of the distance between the noise vectors to that between synthetic samples. The generator pays more attention to the noise vectors under consistent conditions. Moreover, the adversarial loss encourages the generator to probe further into the sample space for similar noise vectors. This also generates samples with large distances in the sample space. Arguably, the distance between samples should be measured from the perspective of the FP network. The FP network's recognition of the samples is reflected by the class probability distribution of synthetic samples. To reduce interference with the generator's learning of the class information, we calculated the JS divergence of the non-target class probability distribution to represent the distance between synthetic samples.

The adversarial loss completes decoupling the noise vector from the condition information. It weakens the generator's attention to the condition. At the same time, it significantly improves the diversity of synthetic samples, which also helps solve the homogenization of intra-class attention.

### 3.2.3. Consistency penalty

The experimental results listed in Table 1 demonstrate that inaccurate BN statistics matching leads to the attention differentiation of synthetic samples in different network modes. Previous studies only concerned the FP network in eval mode. The BN statistical of synthetic samples in training mode failed to match correctly. Therefore, we propose to execute the following consistency penalties on the BN statistical matching loss, classification loss, and coarse-grained attention center matching loss, respectively.

$$Penalty_{BNS}(\boldsymbol{G}) = \sum_{l=1}^{L} \left\| \boldsymbol{\mu}_l^{s,train} - \boldsymbol{\mu}_l \right\|_2^2 + \left\| \boldsymbol{\sigma}_l^{s,train} - \boldsymbol{\sigma}_l \right\|_2^2$$

$$Penalty_{CE}(\boldsymbol{G}) = E_{\mathbf{z},y}(CE(\boldsymbol{F}^{train}(\boldsymbol{G}(\mathbf{z}|y,p)), y))$$

$$Penalty_{CACM}(\boldsymbol{G}) = \max\left(MAE(M^{train}, M^{eval}) - \varepsilon_2, 0\right) \#$$
$$\#(14)$$

where $\varepsilon_2$ represents the relaxing factor.

$Penalty_{BNS}(\boldsymbol{G})$ helps to realize the BN statistics matching under the training mode. $Penalty_{CE}(\boldsymbol{G})$ facilitates that synthetic samples have identical class output in both eval and training modes. $Penalty_{CACM}(\boldsymbol{G})$ reduces the attention matrix difference directly between synthetic samples in different network modes. Remarkably, the attention difference in different modes cannot be corrected effectively merely by $Penalty_{BNS}(\boldsymbol{G})$ because of the massive solution domain of BN statistics matching. Clearly, $Penalty_{CE}(\boldsymbol{G})$ and $Penalty_{CACM}(\boldsymbol{G})$ restricted synthetic samples involving the network behavioral consistency in different modes. This helps the generator realize more reasonable BN statistics matching. In general, the attention differentiation in different network modes of synthetic samples is significantly improved by the interaction between these penalties.

### 3.2.4. Training Process

The quantization training process of ACQ is summarized in Algorithm 1. Like GDFQ, the generator and quantized model of ACQ are trained alternately. To ensure the stable performance of the quantized model, the generator runs warm-up training several times in advance. We use synthetic samples generated in this process to determine the upper and lower bounds of the floating-point number in the quantizer. Relying on the aforementioned loss functions, we can express the overall optimization objective of the generator as follows:

$$L^{ACQ}(G) = L_{CE}^{ACQ}(\boldsymbol{G}) + \epsilon_1 Penalty_{CE}(\boldsymbol{G})$$

$$+\alpha(L_{BNS}^{ACQ}(\boldsymbol{G}) + \epsilon_2 Penalty_{BNS}(\boldsymbol{G}))$$

$$+\beta(L_{CACM}^{ACQ}(\boldsymbol{G}) + \epsilon_3 Penalty_{CACM}(\boldsymbol{G}))$$

$$+\gamma L_{AD}^{ACQ}(\boldsymbol{G})$$
$$\#(15)$$

where $\alpha$, $\beta$, and $\gamma$ represent trade-off parameters; $\epsilon_1$, $\epsilon_2$, and $\epsilon_3$ represent penalty coefficients.

The optimization objective of the quantized model is described as below:

$$L^{ACQ}(\boldsymbol{Q}) = L_{CE}^{ACQ}(\boldsymbol{Q}) + \tau L_{KD}^{ACQ}(\boldsymbol{Q})$$

$$L_{CE}^{ACQ}(\boldsymbol{Q}) = E_{\hat{\boldsymbol{x}}, o^{eval}}\left(CE(\boldsymbol{Q}(\hat{x}), o^{eval})\right) \#(16)$$

where $\tau$ represents the trade-off parameter; $L_{KD}^{ACQ}(\boldsymbol{Q}) = L_{KD}^{GDFQ}(\boldsymbol{Q})$. Unlike GDFQ, the ACQ's generator is decoupled from the quantized model. Not all synthetic samples can be judged as conditional categories by the FP network. For these synthetic samples, the quantized model's output class should align with the judgment result of the FP network to simulate the behavior of the FP network. Therefore, the generator only serves the FP network and is dedicated to synthesizing the required samples. The training of the quantized model is similar to the knowledge distillation. In this process, the cross-entropy loss represents the hard target knowledge of the FP network, whereas the KL distance loss represents the corresponding soft target knowledge.

Algorithm 1: ACQ data-free quantization

**Input**: Pre-trained full-precision model $\boldsymbol{F^{eval}}$, $\boldsymbol{F^{train}}$, warm-up times $T_{warm-up}$, train times $T$.

**Output**: Generator $\boldsymbol{G}$, quantized model $\boldsymbol{Q}$.

Quantize model $\boldsymbol{F^{eval}}$ using Eq (1) and Eq (2), obtain the quantized model Q.

Fix the BN statistics in all BN layers of the quantized model Q.

For $t$ = 1, 2, …, T do

Obtain two groups of random noise $z_1, z_2 \sim N(0, 1)$, class label $y$, and location label $p$.

Obtain two groups of synthetic samples $\hat{x}_1 = G(z_1|y, p)$ and $\hat{x}_2 = G(z_2|y, p)$.

Update Generator $G$ by minimizing Loss $L^{ACQ}(G)$

if $t$ < warm-up times do

execute $Q([\hat{x}_1, \hat{x}_2])$ to obtain the lower bound $l$ and the upper bound $u$

else do

Update the quantized model $Q$ by minimizing loss $L^{ACQ}(Q)$

# 4. Experiment

## 4.1. Dataset and details

We evaluate ACQ on the CIFAR-10, CIFAR-100, and ImageNet2012, which are well-known datasets for evaluating the performance of a model in image classification. CIFAR-10 and CIFAR-100 consist of 50,000 training samples and 10,000 test samples (size: $32 \times 32$). The former contains 10 classes, and the latter contains 100 classes. ImageNet2012 contains1.2 million training samples and 50,000 test samples. The test datasets are only used to evaluate the quantization performance to ensure a data-free application scene in all our experiments.

We quantize Resnet18, MobilenetV2, and Resnet50 based on ImageNet2012, and Resnet-20 based on CIFAR-10 and CIFAR 100. We performed these experiments by using PyTorch based on a hardware platform of NVIDIA RTX 2080Ti GPU. Pytorchcv supplied all our pre-trained models.

We train the generator and the quantized model for 400 epochs (200 iterations/epoch). Specifically, we employed the Adam optimizer to train the generator at an initial learning rate of 0.001. The learning rate is decremented by 0.1 in every 100 epochs. The reflexing factors $\varepsilon_1$ and $\varepsilon_2$ are set at 0.2 and 0.1, respectively, for the hyperparameters of the loss functions of the generator. In ImageNet, we set the penalty coefficients $\epsilon_1$, $\epsilon_2$, and $\epsilon_3$ at 0.05, 0.5, and 1, respectively. We set the trade-off parameters $\alpha$, $\beta$, and $\gamma$ at 0.1, 1, and 0.5 (for MobileNetV2, $\gamma = 0.1$), respectively. In CIFAR-10, We set the penalty coefficients $\epsilon_1$, $\epsilon_2$, and $\epsilon_3$ at 0.5, 1, and 1, respectively. We set the trade-off parameters $\alpha$, $\beta$, and $\gamma$ at 0.5, 1, and 1, respectively. In CIFAR-100, we set the penalty coefficients $\epsilon_1$, $\epsilon_2$, and $\epsilon_3$ at 0.05, 1, and 1, respectively. We set the trade-off parameters $\alpha$, $\beta$, and $\gamma$ at 0.1, 1, and 1, respectively.

We compare ACQ with the existing generative data-free quantization methods and the noise-optimized data-free quantization method. We noted that except for the quality of synthetic samples, the QAT strategy, such as the selection of optimizer and the setting of the hyperparameters (like the training epochs, learning rate, distillation temperature, as well as weight factor $\tau$) has a dramatic impact on the quantization performance. We conduct experiments using the GDFQ training strategy and advanced training strategy AIT, respectively, to compare ACQ with other generative data-free quantization algorithms. In the comparison experiments with the noise-optimized data quantization scheme, we follow the training strategy of IntraQ and present its related experimental results.

## 4.2. Visualization of the ACQ Synthetic Samples

First, we visualize ACQ synthetic samples to verify that the ACQ practically corrects the attention of synthetic samples. In Fig. 6, we provide the same class of synthetic samples generated using Resnet18 and their corresponding attention maps. ACQ synthetic samples have high intra-class diversity in terms of a human's visual perception and network attention compared with GDFQ synthetic samples (Fig. 1(a)). With the attention center position label changing (Fig. 6a), the attention center of synthetic samples become close to the position label. Furthermore, even for identical class labels and attention center position labels (Fig. 6b), due to the coarse-grained matching mechanism and adversarial loss, the synthetic samples and their attention maps are still diverse. We observe from the samples that the ACQ solves the intra-class attention homogenization problem of synthetic samples effectively.

To show the effect of the consistency penalty, we display in Fig. 7 the attention maps of the ACQ synthetic samples in train and eval modes. Although there remained a difference in attention between different modes, compared with the previous GDFQ, the attention similarity of ACQ synthetic samples in the two modes is significantly higher. The ACQ synthetic samples undergo the same experiments as shown in Table 1. The experimental results are listed in Table 2. The BNS error of the ACQ synthetic samples decreased by nearly 60%, and the identification accuracy increased by 32.25% in training mode compared with GDFQ synthetic samples. This observation reveals that the ACQ synthetic samples achieve better BN statistics matching, class consistency, and attention similarity in different modes with consistency penalty.

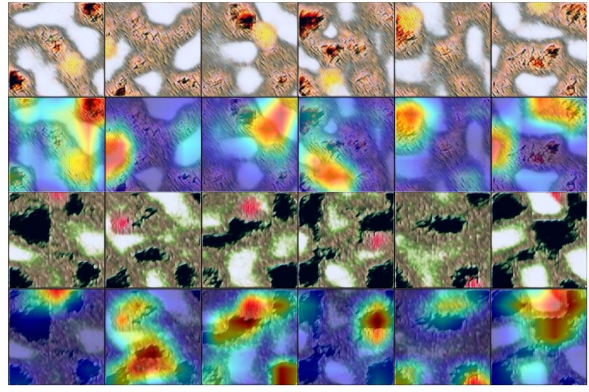

(a)

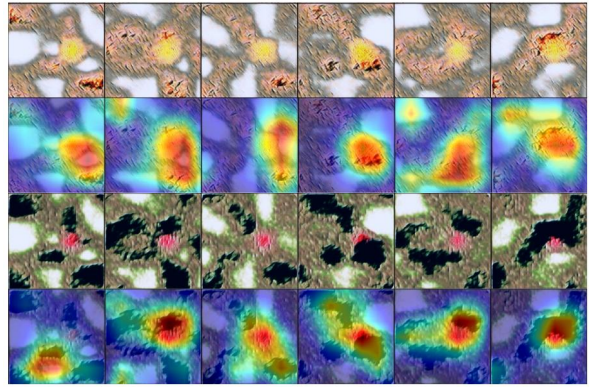

(b)

**Fig.6.** The same class of synthetic data generated by the pre-trained Resnet18 and their attention maps. (a) Identical class labels and different attention center position labels; (b) Identical class labels and attention center position labels.

**Table 2**

The recognition accuracy and BNS error of Resnet18 in different modes based on 960 authentic samples, ACQ synthetic samples, and GDFQ synthetic samples.

|  | Authentic samples | GDFQ synthetic samples | ACQ synthetic samples |
| --- | --- | --- | --- |
| Acc (eval mode) | 100% | 100% | 100% |
| Acc (training mode) | 96.2% | 36.5% | 68.75% |
| BNS error (eval mode) | 0.61 | 0.92 | 1.01 |
| BNS error (training mode) | 0.47 | 2.76 | 1.14 |

Although there is still a huge difference between synthetic samples and real samples from the human perspective, the ACQ effectively improves the problems of homogenization in intra-class attention and attention differentiation in different network modes. We believe that the ACQ synthetic samples are very close to real samples from the perspective of attention and it will improve the performance of the quantized model.

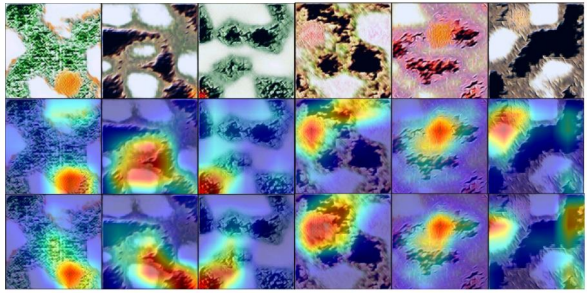

**Fig. 7.** Different classes of synthetic data generated by the pre-trained Resnet18 and their attention maps. The second row of pictures shows the attention maps in eval mode; the third row shows the attention maps in training mode.

## 4.3. Comparison of the Quantization Results

In this section, we present plenty of experiments to demonstrate the outstanding performance of ACQ. In Table 3, we show the classification precisions of ACQ as well as previous generative data-free quantization algorithms with different datasets, objective models, and bit widths employing the training strategy of GDFQ. In this table, $n$w$m$a represents the $n$-bit quantization for weight and the $m$-bit quantization for activation. ACQ improves the performance of the basic framework GDFQ under all settings, especially for large-scale dataset ImageNet with low bit width (4w4a). This observation unravels the significance of correcting the attention of synthetic samples. ACQ surpasses ARC and ZAQ roundly. Although its performance with MobilenetV2 is slightly inferior to that with Qimera, overall, ACQ is more dominant. In particular, for Resnet50, the quantization of ACQ was 2.91% more precise than that of Qimera. This indicates that solving attention problems improves the performance of generative data-free quantization more significantly than other practices.

**Table 3**

Comparison between ACQ and other generative data-free quantization algorithms using the training settings of GDFQ

| Dataset | Model (FP32 Acc %) | Bits | GDFQ | ARC | ZAQ | Qimera | ACQ |
|---|---|---|---|---|---|---|---|
| CIFAR-10 | Resnet20 | 4w4a | 90.25 | 88.55 | 92.13* | 91.26 | 92.07 |
| | 93.89 | 5w5a | 93.38 | 92.88 | 93.36 | 93.46 | 93.55* |
| CIFAR-100 | Resnet20 | 4w4a | 63.39 | 62.76 | 60.42 | 65.10 | 65.61* |
| | 70.33 | 5w5a | 66.12 | 68.40 | 68.70 | 69.02* | 68.76 |

| Dataset | Model | Bits | | | | | |
|---|---|---|---|---|---|---|---|
| | Resnet18 | 4w4a | 60.80 | 61.32 | 52.64 | 63.84 | 64.42* |
| | 71.47 | 5w5a | 68.40 | 68.88 | 64.54 | 69.29 | 69.67* |
| ImageNet2012 | Resnet50 | 4w4a | 52.12 | 64.37 | 53.02 | 66.25 | 69.16* |
| | 77.73 | 5w5a | 71.89 | 74.13 | 73.38 | 75.32 | 76.18* |
| | MobilenetV2 | 4w4a | 59.43 | 60.13 | 0.1 | 61.62* | 61.54 |
| | 73.03 | 5w5a | 68.11 | 68.4 | 62.35 | 70.45* | 69.52 |

* Represents the optimum quantization result.

Moreover, we use the most advanced data-free training strategy, AIT, to show the outstanding performance of ACQ. As illustrated in Table 4, ACQ is superior to other algorithms even though the advanced training strategy helped reduce the performance difference due to the quality of the synthetic images. This is especially the case for ResNet18 and ResNet50. This phenomenon highlights the significance of attention correction. Moreover, ACQ's quantization performance approaches a full-precision model for 5w5a. Specifically, the precision loss due to ACQ was not greater than 1% for Resnet20 (CIFAR-10) and Resnet18. The precision loss was about 2% for Resnet20 (CIFAR-100), Resnet50, and MobilenetV2. As far as we know, this is the best result of data-free quantization.

**Table 4**

Comparison between ACQ and other generative data-free quantization algorithms using the AIT training strategy

| Dataset | Model (FP32 Acc %) | Bits | GDFQ | ARC | Qimera | ACQ |
|---|---|---|---|---|---|---|
| CIFAR-10 | Resnet20 | 4w4a | 91.23 | 90.49 | 91.23 | 91.62* |
| | 93.89 | 5w5a | 93.41 | 92.89 | 93.43 | 93.44* |
| CIFAR-100 | Resnet20 | 4w4a | 65.80 | 61.05 | 65.40 | 65.93* |
| | 70.33 | 5w5a | 69.26* | 68.40 | 69.26 | 68.91 |
| | Resnet18 | 4w4a | 65.51 | 65.73 | 66.83 | 67.55* |
| | 71.47 | 5w5a | 70.01 | 70.28 | 69.22 | 70.66* |
| ImageNet2012 | Resnet50 | 4w4a | 64.24 | 68.27 | 67.63 | 72.23* |
| | 77.73 | 5w5a | 74.23 | 76.00 | 75.54 | 76.29* |
| | MobilenetV2 | 4w4a | 65.39 | 66.47 | 66.81* | 66.81* |
| | 73.03 | 5w5a | 71.11 | 71.96 | 71.68 | 71.98* |

* Represents the optimum quantization result.

Table 5 shows the comparison result between ACQ and the existing noise-optimized data-free quantization algorithms involving the IntraQ training strategy. We observe that ACQ achieves the best result again. A significant performance gap exists between ACQ and the early algorithms ZeroQ and DSG. IntraQ is the most advanced noise-optimized data-free quantization algorithm, which is still inferior to ACQ. ACQ shows clear advantages compared with IntraQ in terms of 4-bit width and CIFAR10/100.

The above experimental results show that ACQ performs best under multiple training settings. This is achieved by solving problems like the homogenization of intra-class attention and attention differentiation in different network modes of synthetic samples. The performance of ACQ under 5-bit width even approaches the performance of an FP model with the most advanced AIT training strategy. ACQ is even superior to the foremost noise-optimized data-free quantization method IntraQ. Accordingly, ACQ could be a better generative data-free quantization method from the perspective of the quality of synthetic samples.

**Table 5**

Comparison between ACQ and noise-optimized data-free quantization algorithms using QAT settings of IntraQ

| Dataset | Model (FP32 Acc %) | Bits | ZeroQ | DSG | GZNQ | IntraQ | ACQ |
|---------|--------------------|------|-------|-----|------|--------|-----|
| CIFAR-10 | Resnet20 | 3w3a | 69.53 | 48.99 | - | 77.7 | 81.98* |
| | 93.89 | 4w4a | 89.66 | 88.93 | 91.30 | 91.49* | 91.16 |
| CIFAR-100 | Resnet20 | 3w3a | 26.35 | 43.42 | - | 48.25 | 51.23* |
| | 70.33 | 4w4a | 63.97 | 62.62 | 64.37 | 64.98 | 66.04* |
| | Resnet18 | 4w4a | 63.38 | 63.11 | 64.50 | 66.47 | 66.81* |
| ImageNet2012 | 71.47 | 5w5a | 69.72 | 69.53 | - | 69.94 | 70.61* |
| | Resnet50 | 4w4a | 60.15 | 60.45 | 53.53 | 65.10 | 65.11* |
| | 77.73 | 5w5a | 70.95 | 70.87 | - | 71.28 | 71.66* |

* Represents the optimum quantization result.

## 4.4. Ablation Experiment

In this section, we conduct ablation experiments to show the effectiveness of coarse-grained attention center matching loss, adversarial loss, and consistency penalty. The experiments are performed on Resnet18, Resnet50, and Resnet20 using the GDFQ training setting under 4-bit width.

In Table 6, we show the impacts of three factors (coarse-grained attention center matching loss, adversarial loss, and consistency penalty) on the network quantization. ACQ is equivalent to GDFQ when the three factors are not used. The Table 6 shows that the coarse-grained attention center matching loss has the most significant impact. In general, the quantization performance increased by more than 2% with the addition of this factor. For Resnet50, the quantization precision even increases by 13.6%. This implies that the homogenization of intra-class attention of synthetic samples has the most decisive negative impact. The adversarial loss has a significant impact on ImageNet2012 compared with the datasets with low spatial dimensions, such as CIFAR-100. This is due to the fact that the mode collapse is likely to happen on the samples with high spatial dimensions. The consistency penalty also enhanced the quantization performance, although the effect of that is inferior to that of coarse-grained attention center matching loss and adversarial loss. As a result, we should pay attention to the consistency in the attention of synthetic samples in different network modes.

**Table 6**

The impacts of the components on the quantization precision

|  | $L_{CACM}$ | $L_{AD}$ | Penalty | result |
|---|---|---|---|---|
|  | × | × | × | 60.80 |
| Resnet18 | √ | × | × | 62.73 |
| (4w4a) | √ | √ | × | 63.61 |
| ImageNet2012 | × | × | √ | 62.42 |
|  | √ | √ | √ | 64.42 |
|  | × | × | × | 52.12 |
| Resnet50 | √ | × | × | 65.72 |
| (4w4a) | √ | √ | × | 67.31 |
| ImageNet2012 | × | × | √ | 61.64 |
|  | √ | √ | √ | 69.16 |
|  | × | × | × | 63.39 |
| Resnet20 | √ | × | × | 65.51 |
| (4w4a) | √ | √ | × | 65.53 |
| CIFAR-100 | × | × | √ | 65.04 |
|  | √ | √ | √ | 65.61 |

The training of the ACQ generator involves a range of hyperparameters. The trade-off coefficient $\gamma$ of the adversarial loss and the penalty coefficient $\epsilon_1$ of the classification loss have a strong impact on the quantization result. In Fig. 7 and Fig. 8, we show their impacts on the quantization performance of the Resnet18. When $\gamma$ increases, the adversarial loss guides the generator to generate diverse samples. This resulted in a marked improvement in the precision of the quantized model. When $\gamma > 0.5$, the adversarial loss took a large proportion of the total loss. This interferes with the generator's learning of the class boundary and attention center, which causes a gradual decrease in the quantization precision. An appropriate penalty coefficient $\epsilon_1$ contributes to improving the judgment consistency of synthetic samples in different modes. However, over large $\epsilon_1$ disperses the generator's attention to other crucial tasks, which could cause poor performance.

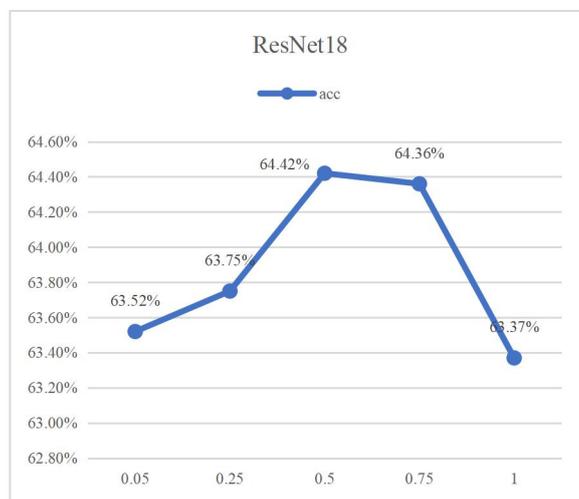

**Fig. 7.** The impact of trade-off coefficient $\gamma$ on the quantization performance.

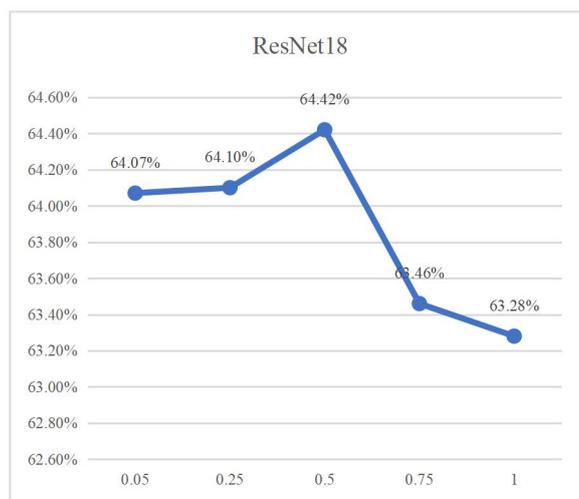

**Fig. 8.** The impact of the penalty coefficient $\epsilon_1$ on the quantization performance.

## 5. Conclusion

In this paper, we revealed the homogenization of intra-class attention and attention differentiation in different network modes of synthetic samples from the perspective of attention. Regarding homogenization, we showed that the quantized model could only learn limited attention modes, whereas the differentiation is caused by inaccurate BN statistical matching and it may mislead the learning of the quantized networks. The performance of generative data-free quantization was hindered by the two problems jointly.

We designed ACQ with an attention center condition generator, given the homogenization of intra-class attention. The attention of synthetic samples became diversified, involving coarse-grained attention matching loss and adversarial loss. The addition of a consistency penalty facilitated synthetic samples to achieve correct BN statistical matching in training mode. It showed identical class judgment and attention to those in eval mode. Accordingly, the differentiation in attention modes was also enhanced.

Given the attention correction of synthetic samples, ACQ displayed the best results with various training strategies in different network quantization experiments using different datasets. The ablation experiment further demonstrated that solving the two abovementioned problems helps to boost the quantization precision significantly.

## Declaration of competing interest

The authors declare that they have no known competing financial interests or personal relationships that could have appeared to influence the work reported in this paper.

## Data availability

Data will be made available on request.

## Acknowledgments

The work is supported by the National Natural Science Foundation of China U1936106, in part by the National Natural Science Foundation of China U19A2080, in part by the CAS Strategic Leading Science and Technology Project XDA27040303, XDA18040400, XDB44000000, in part by the High Technology Project 31513070501 and 1916312ZD00902201. The Beijing Academy of Artificial Intelligence funds this work, and we would like to express our gratitude.